\documentclass[11pt]{article} %11pt

\usepackage{graphicx}
\usepackage{booktabs}
\usepackage{algorithm}
\usepackage{algpseudocode}
\usepackage[T1]{fontenc}
\usepackage{lmodern}
\usepackage[flushleft]{threeparttable}
\usepackage[hidelinks]{hyperref} %
\hypersetup{hidelinks}
\usepackage{multirow}
\usepackage{longtable}
\usepackage[margin=3cm]{geometry} %
\usepackage{amsmath}
\usepackage{authblk}
\usepackage{caption}
\usepackage{cite}

\title{Improving analytical color and texture similarity estimation methods for dataset-agnostic person reidentification}
\author[1,*]{Nikita Gabdullin}
\affil[1]{Joint Stock ``Research and production company ``Kryptonite'' \authorcr
E-mail: n.gabdullin@kryptonite.ru}
\affil[*]{corresponding author}
\date{}
\begin{document}

    \captionsetup[table]{labelformat={default},labelsep=period,name={Table}}

    \maketitle

    \begin{abstract}
        This paper studies a combined person reidentification
        (re-id) method that uses human parsing, analytical feature extraction
        and similarity estimation schemes. One of its prominent features is its
        low computational requirements so it can be implemented on edge devices.
        The method allows direct comparison of specific image regions using
        interpretable features which consist of color and texture channels. It
        is proposed to analyze and compare colors in CIE-Lab color space using
        histogram smoothing for noise reduction. A novel pre-configured latent
        space (LS) supervised autoencoder (SAE) is proposed for texture analysis
        which encodes input textures as LS points. This allows to obtain more
        accurate similarity measures compared to simplistic label comparison.
        The proposed method also does not rely upon persons' photos or other
        re-id data for training, which makes it completely re-id
        dataset-agnostic. The viability of the proposed method is verified by
        computing rank-1, rank-10, and mAP re-id metrics on Market1501 dataset.
        The results are comparable to those of conventional deep learning
        methods and the potential ways to further improve the method are
        discussed.
    \end{abstract}

    
    \section{Introduction}\label{introduction}

Computer vision (CV) is one of the most important areas of computer
science with video analysis playing a crucial role ensuring safety and
security of modern cities. Person reidentification (re-id) is concerned
with identifying same persons captured on different cameras at different
times. Among other things, camera view point and resolution changes,
along with illumination and occlusion effects, make this problem
extremely challenging. Cutting-edge re-id methods primarily rely on deep
learning (DL) which uses large neural networks (NNs) trained on
extremely large datasets. These methods have shown significant success
as illustrated by high benchmark re-id metrics calculated on re-id
datasets~\cite{Rv1,Rv2}. However, they are computationally intensive
which limits their application ``on edge'' in, for instance, smart
cameras. They also suffer from performance decline when transferred from
training setting to real problems ``in the wild''~\cite{GE1,GE2}.

To address these issues, we have previously proposed an alternative
dataset-agnostic method which is not hardware-demanding~\cite{First}. It
combined human parsing~\cite{schpp,preid} performed by a compact neural
network with analytical feature extraction and similarity estimation
schemes. The features included colors analyzed in CIE-Lab (Lab)~\cite{colorsci} color
space and textures analyzed using Local Binary Pattern (LBP)~\cite{lbpp}. The
independence of re-id datasets is ensured by the facts that parser NN is
trained on human parsing datasets~\cite{lipp} and that the analytical part
has no trainable parameters. It has also been shown that the parser can
be sufficiently compact for the complete method to be realized on an
edge device such as Coral TPU with a camera module~\cite{coralreidp}.

Whereas the proposed method has showed promising results by achieving
high rank-1 and rank-10 re-id accuracy metrics comparable to those of DL
methods, its main drawback is low mean average precision (mAP) since it
struggles to correctly rank people that are dressed the same. In this
paper we present several modifications to the original method and study
their effects on re-id metrics. They include smoothing of color channel
histograms and the use of a compact NN with pre-configured latent space
(LS) for texture similarity estimation~\cite{LSc}. Since our previous
work presented a comprehensive comparison of the proposed method with DL
methods on multiple datasets, in this study we only report experimental
results on Market1501 dataset~\cite{Market} and compare them to the ones
obtained in~\cite{First}.

The rest of the paper is organized as follows: Section~\ref{methodology} describes the
methodology, Section~\ref{experiments} summarizes experimental results, and Section~\ref{conclusions}
concludes the paper.

\section{Loss landscape analysis methodology}\label{methodology}

As has already been mentioned in Section~\ref{introduction}, our methodology closely
follows the one previously proposed in~\cite{First}. Therefore, here we
only briefly describe the method and focus on its aspects that are
different from the original. The main purpose is to define several
feature channels that will be used to assess similarity of two persons.

\subsection{Color similarity in Lab color space}\label{colors}

Since human parsing allows us to obtain masks corresponding to different
image regions, color features can be analyzed separately for all regions
to ensure that their attributes are not mixed together. For
every region color information is represented as three histograms
corresponding to color channels. The histograms are converted from RGB
to Lab consisting of one luminosity channel \emph{L} and
two color channels \emph{a} and \emph{b}. This allows to confine all
illumination-related effects to the \emph{L} channel. Histograms are
then compressed from 256 to 64 bins and converted to binary histograms
using a threshold value that depends on the average number of pixels in
each bin. These binary histograms are used to compare same regions
present in a pair of images. The similarity is measured as histogram
intersection normalized by the total number for unique non-zero bins in
both histograms. It can be conveniently expressed as

\begin{equation}
    S_i = \frac{h_{\text{1i}} \ \&  \ h_{\text{2i}}}{h_{\text{1i}} \ \text{||} \ h_{\text{2i}}},
    \label{eq:1}
\end{equation}

where \emph{i} is color channel index that can be \emph{L}, \emph{a}, or
\emph{b}. This approach to histogram similarity calculation 
is more computationally efficient than the
one proposed in~\cite{First}.

\begin{figure} 
    \centering
    \includegraphics[scale=0.55]{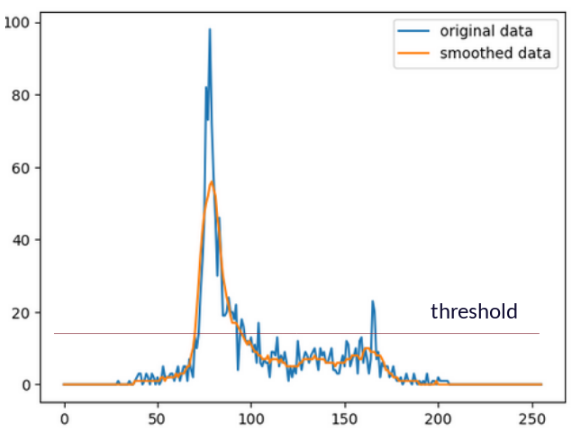}
    \caption{\emph{L} channel histogram before and after smoothing;
    values above the threshold line are used to choose non-zero bins for
    binary histograms.}\label{fig:2-1}
\end{figure}

One drawback of the proposed scheme is that it can be very sensitive to
the noise inevitably present in all histograms, especially \emph{L}
channel ones. Figure~\ref{fig:2-1} shows that this may lead to similar bin values
giving peaks in binary histograms when color intensity oscillates near
the threshold value. To address this issue, it is proposed to perform
histogram smoothing before converting 256 bins into 64 using a
1-dimensional uniform filter. It performs arithmetic averaging for
pixels in \emph{l\textsubscript{f}} consecutive bins along filter
dimension. Figure~\ref{fig:2-1} shows that it allows to reduce noise level and
avoid adding parasitic peaks into binary histograms.

Another similarity measure called \emph{d} is calculated in addition to
color channel similarity measures. It relies on the fact that Lab
color space is uniform, meaning that the Euclidean distance is a true
metric in Lab. This allows color similarity to also be assessed by
checking the distance between a pair of points corresponding to
representative colors of regions in two images. However, distance values
cannot directly be used as similarity measures. Hence, a threshold
hyperparameter is introduced to achieve this, which is inverse to
distance and allows to calculate feature channel \emph{d} similarity
measure. For the exact derivation the reader is referred to Section
2.3.3 in~\cite{First}.

\subsection{Texture similarity with pre-configured LS}\label{textures}

Whereas the original method employed LBP for texture analysis,
this study relies on a compact NN texture classifier. It uses a
Supervised Autoencoder (SAE) with pre-configured LS trained to classify
textures on clothes into five categories: uniform, horizontal lines,
vertical lines, checkered pattern, and dots.

\begin{figure} 
    \centering
    \includegraphics[scale=0.45]{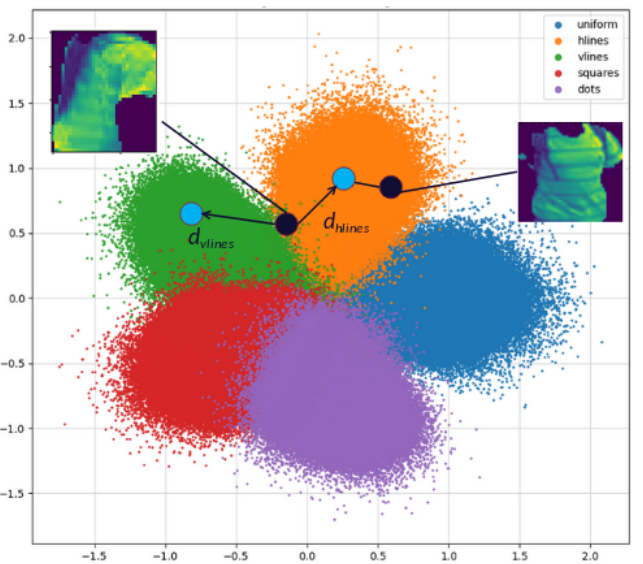}
    \caption{Visualization of pre-configured LS of SAE with projections
    of two images and their distances to centers of the closest clusters.}\label{fig:2-2}
\end{figure}

However, unlike conventional classifiers that just output class labels,
SAE encoder is also trained to embed inputs into specific regions of
2D LS, which are clusters corresponding to texture classes. That is,
encoder output is a position of a specific point in LS which corresponds
to the input texture. This allows a more precise texture comparison to
be performed since re-id problems usually involve analyzing mixed or
obscured textures. Figure~\ref{fig:2-2} shows that texture distances to cluster
centers can be used to calculate texture class similarity vectors. For
the exact derivation the reader is referred to Section 6.1 in~\cite{LSc}.

%fig 2.2

It is important that SAE was not trained on any re-id dataset but on a
custom texture dataset collected from stock images and images
generated with a realistic Stable Diffusion model~\cite{SD}. Therefore,
the addition of texture classifier does not change the dataset-agnostic
property of the proposed method. It also should be noted that texture
similarity estimation is performed only for upper clothes region of the
image, since it often has the most prominent textures.

\subsection{Similarity estimation scheme}\label{sim}

To estimate the similarity of two persons, similarity measures are first
calculated for all parser regions present for both of them. Region
similarity is calculated as weighted sum of similarity measures of all
feature channels

\begin{equation}
    S_c = w_L S_L + w_a S_a + w_b S_b + w_d S_d + w_t S_t,
    \label{eq:2c}
\end{equation}

where \emph{w} are weights of respective feature channels. The total
similarity measure is calculated as a weighted sum of parser region
similarity measures

\begin{equation}
    S_{\text{tot}} = \sum w_c S_c,
    \label{eq:2S}
\end{equation}

where \emph{w\textsubscript{c}} are parser class weights. It should be
noted that whereas \emph{S\textsubscript{c}} is bound to {[}0,1{]}
region, \emph{S\textsubscript{tot}} is not (owing to
\emph{w\textsubscript{c}} not being bound to {[}0,1{]} range, see
Section~\ref{class-exp}) and can be treated more like a similarity score.

\section{Experiments}\label{experiments}

Multiple experiments were conducted to check the viability of
the proposed method and the effect of different parameters on the
results. As previously mentioned, the comprehensive comparison of our
method with conventional DL methods was presented in~\cite{First}. Hence,
here we only compare the modified method with the original one by
calculating the main re-id accuracy metrics rank-1, rank-10, and mAP on
Market1501 dataset.

\subsection{Color histogram modification experiments}\label{color-exp}

This Section studies the effects of \emph{L} channel histogram smoothing
on accuracy of the proposed method. Table~\ref{tab:3-1} shows that histogram
smoothing should be performed before the bin compression, and that
increasing \emph{l\textsubscript{f}} up to 11 allows to improve the
results.

% Table 3.1
\begin{table}
    \caption{The effects of histogram smoothing for \textit{L} color channel on the re-id metrics on Market1501}\label{tab:3-1}
    %\centering
    \begin{tabular}{|l|l|l|l|}
      \hline
        Method & rank-1 & rank-10 & mAP \\
      \hline
      1. No smoothing (base version) & 91 & 96 & 25 \\ \hline
      2. Smoothing after 256-\textgreater64 compression,
      \emph{l\textsubscript{f}} = 11 & 78.7 & 91.2 & 21.2 \\ \hline
      3. Smoothing before 256-\textgreater64 compression,
      \emph{l\textsubscript{f}} = 5 & 91.1 & 95.9 & 24.9 \\ \hline
      4. Smoothing before 256-\textgreater64 compression,
      \emph{l\textsubscript{f}} = 7 & 91 & 96 & 24.7 \\ \hline
      5. Smoothing before 256-\textgreater64 compression,
      \emph{l\textsubscript{f}} = 9 & 91 & 95.7 & 24.4 \\ \hline
      6. Smoothing before 256-\textgreater64 compression,
      \emph{l\textsubscript{f}} = 11 & \textbf{91.9} & \textbf{96} &
      \textbf{25.3} \\ \hline
      7. Smoothing before 256-\textgreater64 compression,
      \emph{l\textsubscript{f}} = 17 & 89.3 & 95.3 & 23.7 \\ \hline
    \end{tabular}
\end{table}

\subsection{Feature channel similarity weight experiments}\label{weights-exp}

This Section studies the effects of feature channel weights in (\ref{eq:2c}) on
the re-id metrics. For classes with no textures other channels' weights
are scaled up while keeping their ratios the same as in Table~\ref{tab:3-2} so
their sum is always 1.

% Table 3.2
\begin{table}
    \caption{The effects of feature channel weights in (\ref{eq:2c}) on the re-id
    metrics on Market1501.}\label{tab:3-2}
    %\centering
    \begin{tabular}{|l|l|l|l|l|l|l|l|l|}
      \hline
      Experiment & \textit{L} & \textit{a} & \textit{b} & \textit{d} & \textit{t} & rank-1 & rank-10 & mAP \\ \hline
      1. Base version & 0.13 & 0.13 & 0.13 & 0.31 & 0.3 & 91 & 96 &
      25 \\ \hline
      2. smaller t & 0.15 & 0.15 & 0.15 & 0.3 & 0.25 & 90 & 96 &
      24.6 \\ \hline
      3. larger t & 0.1 & 0.1 & 0.1 & 0.3 & 0.4 & 87.6 & 95 &
      21.8 \\ \hline
      4. d only & 0 & 0 & 0 & 1 & 0 & 76 & 91 & 18 \\ \hline
      5. no d & 0.24 & 0.23 & 0.23 & 0 & 0.3 & 84.5 & 95 & 19 \\ \hline
      6. no a,b & 0.5 & 0 & 0 & 0.2 & 0.3 & 82.6 & 94 & 15.2 \\ \hline
      7. d \textgreater{} L & 0.2 & 0.1 & 0.1 & 0.3 & 0.3 & \textbf{92.5} & 96
      & \textbf{25.1} \\ \hline
      8. d \textless{} L & 0.3 & 0.1 & 0.1 & 0.2 & 0.3 & \textbf{93.2} & 96 &
      24.3 \\ \hline
      9. large L & 0.25 & 0.15 & 0.15 & 0.15 & 0.3 & \textbf{93.1} & 96 &
      24.2 \\ \hline
      10. low a,b & 0.2 & 0.05 & 0.05 & 0.4 & 0.3 & 92.1 & 96 &
      24.7 \\ \hline
      11. 9 with 6 (Table 3.1) & 0.2 & 0.1 & 0.1 & 0.3 & 0.3 & \textbf{92.9} &
      \textbf{96} & \textbf{25.4} \\ \hline
    \end{tabular}
\end{table}

Table~\ref{tab:3-2} shows that while most modifications make the re-id
metrics worse, they can be slightly improved by increasing \emph{L}
channel weight over \emph{a} and \emph{b} ones. Combining
these weights with the best histogram smoothing parameters obtained in
Section~\ref{color-exp} allows to achieve 1.9\% increase in rank-1 and 0.4\%
increase in mAP, as shown by experiment 11 in Table~\ref{tab:3-2}.

\subsection{Parser class weight experiments}\label{class-exp}

In this Section we study the effects of parser class weights in (\ref{eq:2S}) on
re-id metrics. Table~\ref{tab:3-3} shows that weight modification has not allowed
us to increase re-id metrics over those achieved in~\cite{First}. These
experiments show that the ``upper clothes'' region is indeed the most
important for similarity estimation, though increasing its weight too
much leads to the reduction in accuracy. Neglecting other classes by
reducing their weights also has similar effects.

% Table 3.3
\begin{table}
    \caption{The effects on parser class weights in (\ref{eq:2S}) on the re-id metrics on Market1501}\label{tab:3-3}
    %\centering
    \begin{tabular}{|l|l|l|l|l|l|l|l|l|l|}
      \hline
      Experiment & Upper & Pants & Hair & gloves, & legs & other
      & R-1 & R-10 & mAP \\
      & clothes & & & boots & & & & & \\
      \hline
      1. base & 8 & 6 & 3 & 2 & 1 & 1 & \textbf{91} & \textbf{96} &
      \textbf{25} \\ \hline
      2. & 10 & 6 & 3 & 2 & 1 & 1 & 88.3 & 95.2 & 22.1 \\ \hline
      3. & 8 & 4 & 1 & 1 & 1 & 1 & 90.6 & 96 & 24.4 \\ \hline
      4. & 8 & 2 & 1 & 1 & 1 & 1 & 89.5 & 95 & 22.7 \\ \hline
      5. & 6 & 2 & 1 & 1 & 1 & 1 & 88.6 & 94.9 & 22.4 \\ \hline
      6. & 1 & 1 & 1 & 1 & 1 & 1 & 81.8 & 90.8 & 16.4 \\ \hline
    \end{tabular}
\end{table}

\subsection{Gallery search without query images}\label{noimg}

One interesting feature of the proposed method is the possibility to
conduct gallery search without query images, relying only on textual or
verbal description of a person. Unlike conventional NN feature vectors,
all features described in Section~\ref{methodology} are interpretable and have
human-readable format. Therefore, one could construct feature vectors
corresponding to some person's description even if their image is not
available. The method for constructing color feature vectors was
previously described in Section 4.3 in~\cite{First}, and its texture
counterpart was described in Section 6.3 in~\cite{LSc}. In this Section
we combine these two methods to perform Market1501 search using only
textual descriptions.

\begin{figure} 
  \centering
  \includegraphics[scale=0.44]{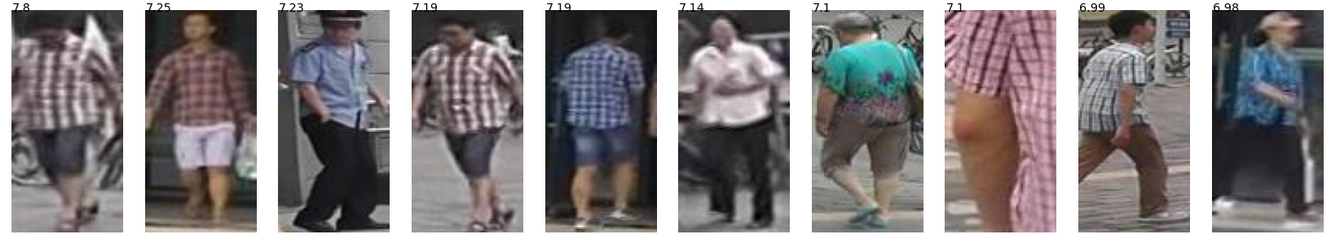}
  \caption{Search results for persons wearing ``checkered upper clothes''
  in Market1501. Numbers above images show similarity score out of 8.}\label{fig:3.4-1}
\end{figure}

Firstly, Figure~\ref{fig:3.4-1} shows top-10 search results for persons wearing
``checkered upper clothes'' regardless of its color. Since only texture
description is provided in this query, the weight of texture feature
channel is set to 1 for this search, and no other region but upper
clothes is used for similarity score calculation. Apart from examples
where poor quality, folds, and shirt image print confuses SAE, all
images indeed show different people wearing checkered shirts.

\begin{figure} 
  \centering
  \includegraphics[scale=0.44]{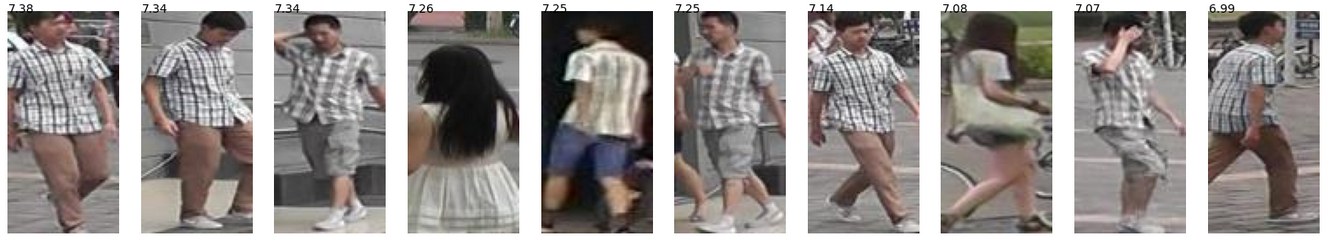}
  \caption{Search results for persons wearing ``white checkered upper
  clothes'' in Market1501. Numbers above images show similarity score out
  of 8.}\label{fig:3.4-2}
\end{figure}

Secondly, the previous query can be made more specific by, for instance,
specifying white color in upper clothes' description. Since both color
and texture feature channels are used in this example, feature channel
weights shown in Table~\ref{tab:3-2} are used to conduct this search. In this case
different people wearing white checkered shirts can indeed be found, as
shown in Figure~\ref{fig:3.4-2}. There are two false results of women wearing
plain clothes with uniform textures obscured by hair in one case and a
bag in another one. Nevertheless, the method overall succeeds in
providing consistent results that are very close to the query
description.

\begin{figure} 
  \centering
  \includegraphics[scale=0.44]{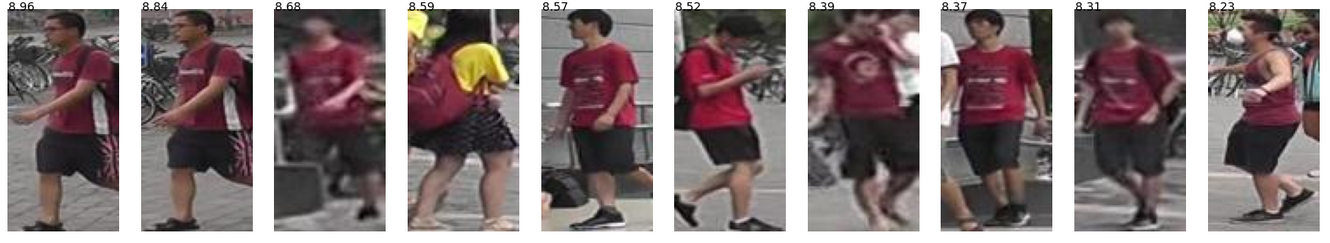}
  \caption{Search results for persons wearing ``red upper clothes and
  black pants'' in Market1501. Numbers above images show similarity score
  out of 14.}\label{fig:3.4-3}
\end{figure}

Finally, search results for persons wearing ``red upper clothes and
black pants'' are shown in Figure~\ref{fig:3.4-3}. Since multiple clothes
corresponding to different parser classes are described in the query,
feature vectors are constructed separately for each class and class
weights described in Table~\ref{tab:3-3} are used for the analysis. Figure~\ref{fig:3.4-3}
shows only one false result, which is a consequence of the parser
incorrectly labeling the backpack as ``upper clothes'' and ignoring the
yellow shirt completely. Different shirt images can be explained by SAE
not being trained to distinguish specific prints, resulting in their
close proximity in LS.

\subsection{Discussions}\label{discussions}

The results in this Section show that the proposed modifications result
only in a slight increase in the re-id metrics on Market1501. One of the
reasons why the improved texture comparison methodology cannot
significantly contribute to the re-id accuracy is that Market1501 has
very limited texture variability. That is, this dataset does not contain
clothes with prominent textures apart from a small number of striped
shirts and dresses. Furthermore, many different persons are dressed in
very similar clothes such as, for instance, University uniform shirts
and shorts.

The results shown in this Section allow to conclude that further
improvement of color and texture features cannot improve the overall
performance of the method. Moreover, varying some parameters often leads
to worse results. This means that new features that are qualitatively
different from the colors and textures are required. The most promising are features that could account
for person's height and physique. These features would allow to rank
similarly dressed persons without relying on color and texture
information. Unfortunately, such features are hard to formulate
analytically due to their extreme sensitivity to perspective changes due
to variations in camera view angle and person's pose. It is possible to
extract these features using NNs, though this would further increase the
computational burden of the method. Such features are also unlikely to
be interpretable complicating the search technique described in Section~\ref{noimg}.

This paper shows that dataset-agnostic SAE texture NN for similarity
estimation can indeed be used for re-id purposes. The proposed scheme
that uses image embeddings' LS positions for similarity estimation is
also very convenient since it essentially outputs a percentage
similarity estimate rather than a label chosen from a predefined class
list. This is useful when analyzing partly obscured or noisy textures
such as clothes with folds or shadows. Since LS position encodes
textures in human-readable format, it allows non-ambiguous conversion of
verbal texture descriptions directly into LS embeddings for similarity
search. SAE discriminative ability can be
further improved using more diverse texture classes during
training. Future work will also investigate other types of LS
configuration to further improve generalization capabilities of NNs.

\section{Conclusions}\label{conclusions}

This paper presents a dataset-agnostic approach to person
reidentification that is not computationally demanding and can be
implemented on edge devices such as smart cameras. It combines human
parsing, used to identify important image regions, with analytical
feature extraction and similarity estimation scheme for their analysis.
The proposed features include color and texture channels each compared
independently for same regions in a pair of images. It is proposed to
analyze color features in Lab color space and use histogram smoothing
for noise reduction. Texture similarity evaluation is facilitated by SAE
with pre-configured LS which allows to get a more accurate similarity
measure by directly comparing embeddings corresponding to two textures.
Unlike DL re-id NNs, SAE is trained on texture dataset, which makes it
re-id dataset-agnostic and ensures good generalization. The
applicability of the proposed method is illustrated by calculating re-id
accuracy metrics on Market1501 that shows results comparable with DL
methods. The experimental results also show that metrics cannot be
improved further by additional research in color and texture features,
and new features that describe other aspects of a person are required.

\section*{Acknowledgement}\label{acknowledgement}

The author would like to thank his colleagues Dr Anton Raskovalov, Dr
Igor Netay, and Ilya Androsov for fruitful discussions, and Vasily
Dolmatov for discussions and project supervision.

\bibliographystyle{IEEEtran}
\bibliography{IEEEabrv,ms}

\end{document}